# Analyzing decision tree bias towards the minority class

Nathan Phelps[a]*, Daniel J. Lizotte[bc†], and Douglas G. Woolford[a†]

*a. Dept. of Statistical and Actuarial Sciences, University of Western Ontario, London, Canada;*
*b. Dept. of Computer Science, University of Western Ontario, London, Canada;*
*c. Dept. of Epidemiology and Biostatistics, University of Western Ontario, London, Canada*

*Corresponding author: Email: nphelps3@uwo.ca

[†]Equal contribution

ORCID for Nathan Phelps: 0000-0002-3173-3368

ORCID for Daniel J. Lizotte: 0000-0002-9258-8619

**Abstract:** There is a widespread and longstanding belief that machine learning models are biased towards the majority class when learning from imbalanced binary response data, leading them to neglect or ignore the minority class. Motivated by a recent simulation study that found that decision trees can be biased towards the minority class, our paper aims to reconcile the conflict between that study and other published works. First, we critically evaluate past literature on this problem, finding that failing to consider the conditional distribution of the outcome given the predictors has led to incorrect conclusions about the bias in decision trees. We then show that, under specific conditions, decision trees fit to purity are biased towards the minority class, debunking the belief that decision trees are always biased towards the majority class. This bias can be reduced by adjusting the tree-fitting process to include regularization methods like pruning and setting a maximum tree depth, and/or by using post-hoc calibration methods. Our findings have implications on the use of popular tree-based models, such as random forests. Although random forests are often composed of decision trees fit to purity, our work adds to recent literature indicating that this may not be the best approach.

**Statements and declarations:** The authors do not have any relevant financial or non-financial interests to disclose.

**Acknowledgements:** We acknowledge the support of the Natural Sciences and Engineering Research Council of Canada (NSERC) through its Postgraduate Scholarship program, Discovery Grant program, and Strategic Networks program. We also acknowledge financial support provided through an Ontario Graduate Scholarship, as well as assistance from ChatGPT, which performed tasks such as brainstorming, conducting initial simulations, and helping develop proofs.

## 1. Introduction

There are several very important fields in which difficult imbalanced binary classification problems occur. These include the prediction of cancer (e.g., Fotouhi et al., 2019), flooding (e.g., Tanimoto et al., 2022), suicidal ideation (e.g., Ben Hassine et al., 2022), and terrorism (e.g., Zheng et al., 2022). In such cases, one of the two classes occurs much less frequently than the other. This class is typically known as the minority or positive class and is generally denoted by 1; the other class is typically called either the majority or negative class and is denoted by 0 (or sometimes -1). In the machine learning/artificial intelligence community, there is a widespread and longstanding belief that machine learning models perform poorly on such data due to a bias towards the majority class (e.g., Japkowicz and Stephen, 2002; Guo et al., 2008; Leevy et al., 2018; Megahed et al., 2021). This could manifest either as a classifier outputting class predictions that are disproportionately the majority class or as a model outputting probability estimates that are biased towards 0. In either case, this is problematic because models that "neglect" (Japkowicz and Stephen, 2002) or "ignore" (Guo et al., 2008) one of the two classes, especially the class we are typically most interested in, cannot be relied upon in practice.

Several methods have been developed to try to reduce or eliminate this anticipated bias. These methods generally involve either preprocessing the data through sampling techniques, such as under- or over-sampling (e.g., Megahed et al., 2021) and the Synthetic Minority Oversampling Technique (SMOTE) (Chawla et al., 2002), or adjusting the machine learning algorithm through cost-sensitive learning (e.g., Chen et al., 2004; Sun et al., 2007; Krawczyk et al., 2014). Those methods themselves can lead to creating poorly calibrated models (i.e., their predictions do not reasonably represent event probabilities), so other methods have been employed to account for using these approaches, such as analytical calibration (e.g., Dal Pozzolo et al., 2015), Platt's scaling (Platt, 1999), and isotonic regression (e.g., Zadrozny and Elkan, 2002).

However, in the case of decision trees, and possibly other machine learning models based on them (e.g., random forests), it may be that all this work has been done without proper justification. To our knowledge, decision trees have never been excluded from the group of machine learning models thought to underestimate or ignore the minority class; some studies on class imbalance have even had a special emphasis on decision trees (e.g., Japkowicz and Stephen, 2002). However, recent work has provided evidence that decision trees can actually be biased towards the *minority* class (Phelps et al., 2024). In that study, decision trees fit to purity (i.e., perfect separation of positive and negative cases)—which is common practice when using decision trees to create a random forest (Zhou and Mentch, 2023)—did not neglect the minority class; rather, they systematically *overestimated* the proportion of observations belonging to the minority class. This contradiction to a seemingly universally held belief has revealed a lack of understanding regarding the bias in decision trees in the context of imbalanced binary classification.

Motivated by these recent findings, here we provide new theoretical evidence explaining why decision trees can be biased towards the minority class and present strategies for reducing this bias. We begin with a critical review of past literature which is followed by analyses of illustrative cases that demonstrate the mechanism for bias. Finally, we discuss the implications of our work and outline several avenues for future work.

## 2. Literature review

Going back two decades, there are many studies that address the class imbalance problem, with claims including that machine learning models "underestimate", "ignore", or "neglect" the minority class (e.g., Japkowicz and Stephen, 2002; Guo et al., 2008; Megahed et al., 2021). This is a core problem in the machine learning community, with a large body of work devoted to it. See, for example, the reviews of Leevy et al. (2018) and Rezvani and Wang (2023) for detailed summaries of the vast literature on this topic. Decision trees are one of the machine learning models that have been criticized for their performance on imbalanced data (e.g., Japkowicz and Stephen, 2002). However, in a recent simulation study that considered varying levels of class imbalance in the data, Phelps et al. (2024) found that their decision trees tended to overpredict the number of positive cases. The overestimation generally increased as the level of class imbalance increased and, in some cases, led to predicting more than 10% more positive cases than were present in the data. These results provide us with reason to revisit the longstanding belief that decision trees are biased towards the majority class.

The criticism of the performance of decision trees on imbalanced data has led to a number of studies being conducted to improve upon the traditional decision tree algorithm in this context (e.g., Cieslak and Chawla, 2008; Prati et al., 2008; Liu et al., 2010; Boonchuay et al., 2017). Some of these, however, have focused on improving decision trees with respect to area under the receiver operating characteristic curve (AUC), which is different from the focus of our study. AUC addresses the ranking of the observations in terms of their likelihood of being a positive case, not under- or over-prediction with respect to the true outcomes, so we do not focus on those studies. In our literature review, we pay special attention to two papers that have shown decision trees are biased towards the majority class, one that has shown this for decision trees that output class predictions (Japkowicz and Stephen, 2002) and one that has shown this for decision trees that output class probabilities (Wallace and Dahabreh, 2014).

In one of the earliest studies of the class imbalance problem, Japkowicz and Stephen (2002) reported that C5.0 decision trees "neglect" the minority class. In their study, decision trees were not fit to purity and were used to make class predictions. Although not explicitly discussed, this means that the models generated scores, based on the proportion of positive cases in their leaf nodes, for each observation on which they made a prediction. Oftentimes, these scores are treated as estimates of the probability of belonging to the positive class. To generate class predictions, the scores are mapped to 0 or 1 according to a decision threshold. As noted by Collell et al. (2018) and Esposito et al. (2021), this threshold is commonly set to 0.5. Japkowicz and Stephen (2002) did not specify their threshold, so we assume they followed this convention. However, a threshold of 0.5 may not be sensible when modeling imbalanced data. This becomes clear through critical consideration of the conditional distribution of the outcome given the predictors; it is entirely plausible to have a conditional distribution whose probabilities never exceed 0.5. This may be particularly relevant when predicting the occurrence of rare events. For example, consider the data generating process from the simulation study in Phelps et al. (2025), where the mean probability of success is approximately 0.0022. In Fig. 4 of that paper, none of the one million observations have a probability of being a positive case that exceeds 0.06. Thus, even if the scores output by the decision trees perfectly estimate the probability of being a positive case given the predictors, it is correct to classify every observation as a negative case when using a threshold of 0.5. Of course, such a model is not useful. However, this should not be treated as evidence of a problem with decision trees; the problem in this case is the decision threshold. This argument casts doubt on findings of a bias towards the majority class that are based on decision trees that classify data based on a threshold of 0.5. We are not aware of any studies that have struggled with ignoring the minority class when using a more appropriate threshold to account for class imbalance, and multiple studies have found success when doing so (e.g., Collell et al., 2018; Esposito et al., 2021).

We have shown that decision trees could appear biased towards the majority class because of the decision threshold used, even when the decision tree's scores perfectly estimate the probability of a case being positive. However, that argument says nothing about whether the scores themselves are unbiased estimates of these probabilities. This aspect also needs to be addressed, as probability estimates from tree-based models have also been criticized in the literature. Using multiple machine learning models, including boosted decision trees, Wallace and Dahabreh (2014) showed that "probability estimates obtained via supervised learning in imbalanced scenarios systematically underestimate the probabilities for minority class instances", describing them as "unreliable". However, their statements were largely based on observing that estimates for minority class observations were small, as opposed to *too small*. This, again, does not consider the conditional distribution of the outcome given the predictors. Recall the simulation study in Phelps et al. (2025) where none of the one million observations had a probability of being a positive case that was larger than 0.06. With a model that perfectly estimates the true probabilities, one would still obtain results like those in Wallace and Dahabreh (2014); predictions for the majority class will be good, but predictions for the minority class may seem bad, even though they are perfect. Such estimates have been unfairly classified as "unreliable" simply because they are small. Wallace and Dahabreh (2014) theoretically justify their findings by pointing to the bias in logistic regression (King and Zheng, 2001), but this bias does not account for the small probability estimates attributed to minority class observations. Consider the special case they discuss, where the bias in the estimate of $\beta_0$ is $\mathbb{E}\left[\hat{\beta}_0 - \beta_0\right] \approx \frac{\tilde{\pi} - 0.5}{n\tilde{\pi}(1-\tilde{\pi})}$. Here, $\tilde{\pi}$ is the average of success probabilities for observations in the dataset—which for large datasets can reasonably be approximated with the prevalence of the minority class—and $n$ is the total number of observations in the dataset. Consider, for example, a sample of 500 observations from a data generating process with a true prevalence of 2%. In this case, $\mathbb{E}[\hat{\beta}_0 - \beta_0] \approx -0.049$. Note that this bias is on the log-odds scale. Thus, it is most impactful on probability estimates when they are near 0.5, and even then, an estimate that should have been 0.5 is reduced only to 0.488. While there is a bias, it is not substantially changing the estimated probabilities of belonging to the minority class. Additionally, this bias was derived only for logistic regression, not other models.

Our preceding literature review suggests that the belief that decision trees are biased towards the majority class is not well-founded, especially considering the bias towards the minority class observed in Phelps et al. (2024). Like Phelps et al. (2024), Plante and Radatz (2024) also numerically investigated the biases of tree-based models, finding biases (some of which were small enough to potentially just be due to chance) in either direction depending on the model and dataset. Neither study provided any theoretical justification of a bias in decision trees. Considering the mixed results in the literature, we believe additional theoretical analyses are needed to provide a more solid foundation for understanding the bias in decision trees. The present study addresses this need.

## 3. Investigating the bias under different scenarios

We present two different scenarios—one where the outcome is deterministic given a predictor and one where the outcome is stochastic and unrelated to the predictors—and consider the bias in decision trees fit to purity under each setting. In both, we show that there are configurations that lead to decision trees being biased towards the minority class.

### 3.1 A deterministic case with full information

Consider a simple case where we have only one covariate and an outcome that is completely determined by that covariate. Since the outcome is deterministic, only one split is needed to separate the positive and negative outcomes. Through constructing such a scenario, we can compare the threshold learned by the decision tree, $T'$, to the true threshold, $t$, to determine how $T'$ might be a biased estimate of $t$.

Consider the following data generating process. We have a random variable, $Z \sim \text{Bernoulli}(p)$, where $p \in (0,1)$, and another random variable, $X$, which is the covariate used for modeling and comes from the following distribution:

$$X \sim \begin{cases} \text{Unif}(0, t) \text{ if } z = 0 \\ \text{Unif}(t, 1) \text{ if } z = 1 \end{cases}$$

In this situation, $t \in (0, 1)$. The deterministic outcome, $Y$, is determined as indicated below:

$$Y = \begin{cases} 0 \text{ if } x < t \\ 1 \text{ if } x \geq t \end{cases}$$

To create a dataset for modelling, we draw from this data generating process $n$ times. An illustration of the process is provided in Fig. 1.

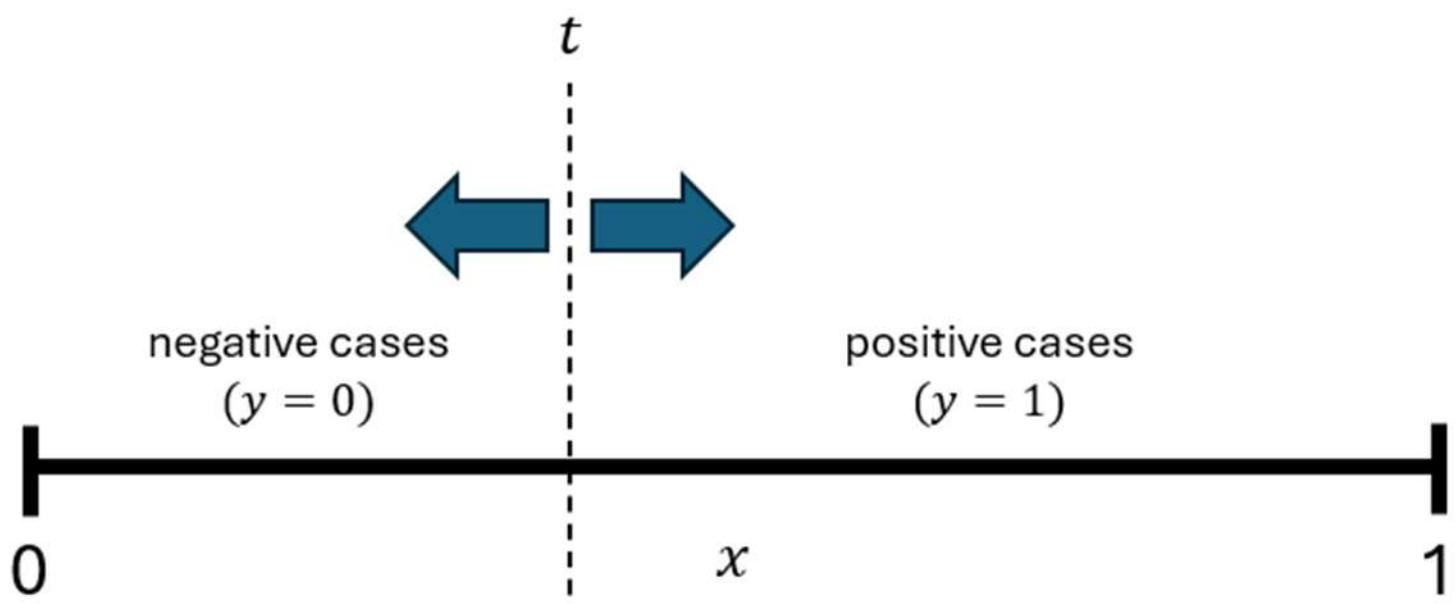

**Fig. 1** An illustration of the data generating process for the deterministic case with full information

Consider training a decision tree on such a dataset. Assuming the dataset contains at least one observation from each class, the decision tree will learn a threshold equal to the average of $\max(X_i|X_i < t)$ and $\min(X_i|X_i \geq t)$, where $i = 1, 2, \ldots, n$. Since $X|X < t$ and $X|X \geq t$ both follow uniform distributions, it is straightforward to compute the expected maximum and minimum. Let $n_{\text{pos}}$ represent the number of positive observations in the dataset. For $k \in 1, 2, \ldots, n-1$, we can then obtain the following:

$$\begin{aligned}\mathbb{E}\left[T'\middle|n_{\text{pos}} = k\right] &= \frac{1}{2}[\max(X_i|X_i < t) + \min(X_i|X_i \geq t)] \\ &= \frac{1}{2}\left[\left(\frac{t(n-k)}{n-k+1}\right) + \left(t + \frac{1-t}{k+1}\right)\right]\end{aligned} \tag{1}$$

For the degenerate cases where the dataset is composed of entirely positive or entirely negative outcomes, we assign thresholds of 0 and 1, respectively, retaining the usual format where positive outcomes are to the right of the threshold and negative outcomes are to the left of the threshold. Then, by considering the probability of $n_{\text{pos}}$ positive observations, we can compute $\mathbb{E}[T']$ as follows:

$$\mathbb{E}[T'] = \sum_{k=1}^{n-1} \binom{n}{k} p^k (1-p)^{n-k} \mathbb{E}[T'|n_{\text{pos}} = k] + (1-p)^n \tag{2}$$

Thus, given $n$, $p$, and $t$ of a data generating process, we can compute $\mathbb{E}[T']$. Consider an example where $n = 100$, $p = 0.05$, and we allow $t$ to vary from $0.001$ to $0.999$ in increments of $0.001$. The two plots in Fig. 2 compare $\mathbb{E}[T']$ and $t$, with the solid line showing the relationship between them and the dashed line providing a reference for the line that exists if $\mathbb{E}[T'] = t$ (i.e., $T'$ is an unbiased estimator for $t$). Upon a visual inspection, it seems as though $\mathbb{E}[T']$ is always at least as large as $t$. This indicates a bias towards the majority class, since $\mathbb{E}[T']$ being too big shrinks the size of the region attributed to the positive class, which is our minority class.

However, although it is not visible in the plots, $\mathbb{E}[T']$ is actually smaller than $t$ when $t$ is sufficiently large. This is better illustrated in the next example.

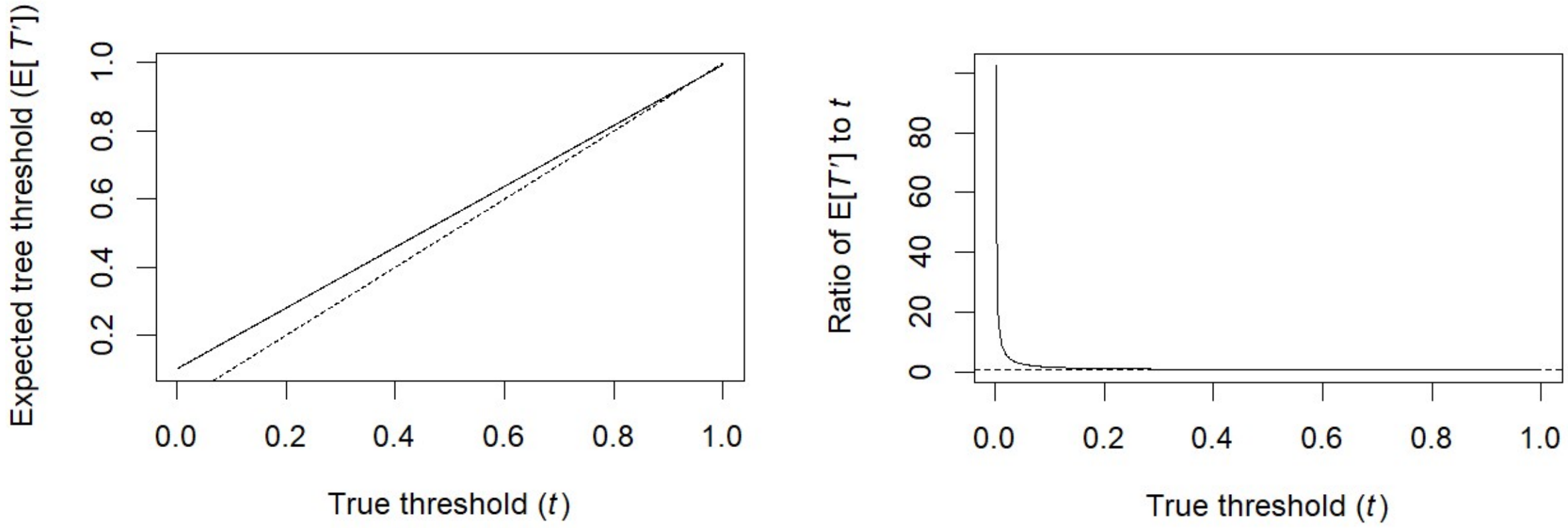

**Fig. 2** Plots comparing the expected threshold learned by a decision tree to the true threshold in the data generating process. The solid line shows the relationship between them and the dashed line provides a reference for the line that would exist if the two were equal. In this data generating process, 5% of observations belong to the positive class. The datasets have 100 observations

Now consider the case where $p = 0.95$, leaving the remaining settings the same as before. Note that this results in the positive cases being the majority class, differing from the usual convention. The plots in Fig. 3 show that such a set-up can also lead to biases. Here, the left plot appears to show that $\mathbb{E}[T']$ is always at least as small as $t$, which would again indicate a bias towards the majority class because the minority class is the negative cases, which occur to the left of the threshold. However, the right plot shows clearly that the ratio of $\mathbb{E}[T']$ to $t$ is below 1 for most values of $t$, but also above 1 for sufficiently small $t$. These examples clearly show that decision trees can be biased in either direction; depending on the data generating process, a decision tree can be biased towards the majority class or the minority class. Therefore, one should not always assume that a decision tree will be biased towards the majority class when it is used for an imbalanced classification problem.

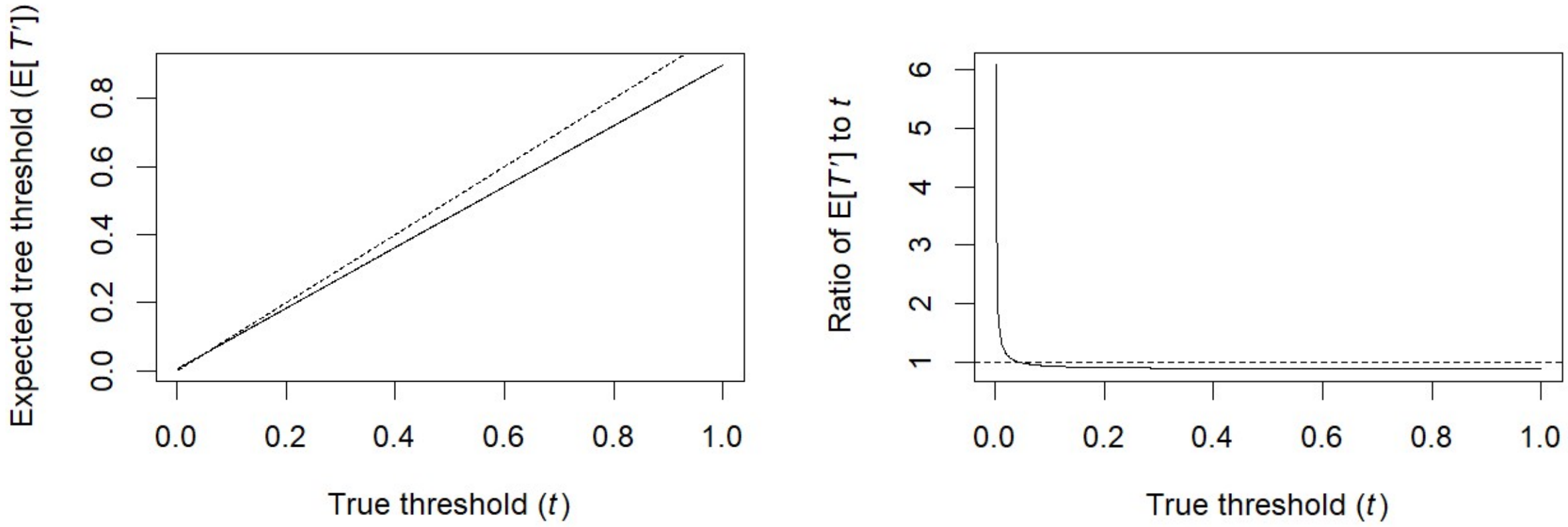

**Fig. 3** Plots comparing the expected threshold learned by a decision tree to the true threshold in the data generating process. The solid line shows the relationship between them and the dashed line provides a reference for the line that would exist if the two were equal. In this data generating process, 95% of observations belong to the positive class. The datasets have 100 observations

## 3.2 The entirely stochastic case

Now consider a regime where the data generating process involves no relationship between the covariates and the outcome (i.e., entirely stochastic). Like before, we have datasets of $n$ observations and an outcome, $Y$, that takes values 0 or 1, but we now generalize to having $m$ covariates. We consider decision trees that are fit to purity, as this is commonly done when fitting decision trees as part of a random forest (Zhou and Mentch, 2023). This was also the case in Section 3.1, but only one split was needed to achieve this. Unlike in Section 3.1, this environment does not have a true threshold where the probability of the outcome being 0 or 1 changes as the threshold is crossed. Thus, we cannot consider the bias in decision trees in the same way. Instead, we will now consider prevalence estimates relative to the true prevalence of the positive class. We define the prevalence estimate, $P_E$, of a tree to be its expected output with respect to the joint distribution of the covariates. More formally, with $A \subseteq \mathbb{R}^m$ representing the subset of the predictor space for which the tree predicts a positive case and $\mathbb{P}_{\boldsymbol{X}}$ representing the joint distribution of the predictors, we define $P_E \equiv \int_A \mathbf{1}\, d\mathbb{P}_{\boldsymbol{X}}(\boldsymbol{x}) = \mathbb{P}(\boldsymbol{X} \in A)$. Note that this definition requires knowledge of $A$, but $A$ is unknown until a tree is fit. Thus, the expected prevalence estimate of a tree refers to the expectation over the distribution of $A$, which depends on the data generating process and the tree-fitting procedure.

### *3.2.1 Additional assumptions*

Throughout this section, we assume that the predictors are uniformly distributed, which substantially simplifies our computations. Initially, it might seem that this assumption will limit our findings so much so that they would be useless in practice. However, any continuous distribution can be converted to a $\text{Unif}(0,1)$ distribution using the probability integral transform (Angus, 1994)[1]. In addition, when a split is performed on a predictor, the predictor still follows a uniform distribution on either side of the split, just with different ranges. Thus, if we can understand the bias in decision trees when predictors are uniformly distributed, we can understand the bias in decision trees more generally by first transforming the distributions of each predictor.

We also assume that the decision tree's splits are all based on the same covariate, chosen by the decision tree when it makes its first split. This differs from the decision tree algorithms used in practice, but we will show that this assumption is not the cause of the biases we find.

*3.2.2 A special case: One positive observation*

We start by considering the special case where we have only one positive observation in the dataset. This is not a realistic setting in practice, as having only one positive case would be prohibitive for meaningfully training a decision tree. However, it provides a simpler setting for developing a deeper understanding of the bias in decision trees.

**Theorem 1** *Consider a data generating process that produces datasets of* $n$ $(n > 2)$ *instances, each with* $m$ $(m \geq 2)$ *independent predictors and a label that is 0 or 1. Thus, the dataset of predictors,* $\boldsymbol{X}$*, is* $n \times m$*. Within each dataset, one instance is uniformly randomly assigned label 1, irrespective of the covariate values. Provided* $X_j \sim Unif(0, u_j), u_j > 0, \forall\, j \in 1, 2, \ldots, m$*, then a decision tree fit to purity based on only one of the* $m$ *predictors (chosen by the decision tree algorithm) produces an expected prevalence estimate (*$\mathbb{E}[P_E]$*) of* $\left(\frac{1}{2}\right)\left(\frac{1}{n+1}\right)\left[3 - \left(\frac{n-2}{n}\right)^m\right] > \frac{1}{n}$*. Thus, the decision tree provides an estimate biased towards the minority class.*

**Proof:** Without loss of generality, we assume $u_j = 1\ \forall j \in 1, 2, \ldots, m$. This allows us to compute $P_E$ by considering just the size of the part of the tree that predicts a label of 1, without need for normalization. Further, since we are considering a decision tree that splits on only one predictor, we can represent the size of the part of the tree that predicts a label of 1 using the length of the region of this predictor that predicts a label of 1. Thus, $P_E$ is based only on the length of this region. Let $\boldsymbol{X}_s$ be the vector of covariate values from $\boldsymbol{X}$ for the predictor chosen by the decision tree and let $S_1, S_2, \ldots, S_{n+1}$ be the spacings on $[0,1]$ between the ordered realizations of $\boldsymbol{X}_s$, including the boundaries. We also let $K$ represent the index of the positive case when the realizations of $\boldsymbol{X}_s$ are sorted in increasing order. Decision trees make splits halfway between the observations, so if $K \in (2, 3, \ldots, n-1)$, then $P_E = \frac{1}{2}(S_K + S_{K+1})$. If $K = 1$, then $P_E = S_1 + \frac{1}{2}S_2$. Likewise, if $K = n$, then $P_E = S_{n+1} + \frac{1}{2}S_n$. We can then construct the following equation for $P_E$:

[1] Strictly speaking, using the probability integral transform requires knowing the original distribution, which will generally not be the case. However, we can still convert any distribution to approximately $\text{Unif}(0,1)$.

$$
\begin{aligned}
P_E &= \frac{1}{2}\sum_{i=2}^{n-1} I(K=i)(S_i + S_{i+1}) + I(K=1)\left(S_1 + \frac{1}{2}S_2\right) + I(K=n)\left(S_{n+1} + \frac{1}{2}S_n\right) \\
&= \frac{1}{2}\sum_{i=1}^{n} I(K=i)(S_i + S_{i+1}) + \frac{1}{2}[I(K=1)S_1 + I(K=n)S_{n+1}]
\end{aligned} \tag{3}
$$

Our goal is to compute $\mathbb{E}[P_E]$. Since $\boldsymbol{X}_s$ consists of independent and identically distributed draws from Unif(0,1), it is straightforward to compute $\mathbb{E}[S_i] = \frac{1}{n+1}$ for all $i$. All that remains then is to compute $\mathbb{P}(K = i)$, but we only need to compute the probability that $K \in (1, n)$. This is also straightforward to compute because the decision tree will always choose to split on a predictor where $K \in (1, n)$ if such a predictor exists. Thus, $\mathbb{P}\big(K \notin (1, n)\big) = \left(\frac{n-2}{n}\right)^m$ and therefore $\mathbb{P}\big(K \in (1, n)\big) = 1 - \left(\frac{n-2}{n}\right)^m$, so we obtain:

$$
\begin{aligned}
\mathbb{E}[P_E] &= \frac{1}{2}\left(\frac{2}{n+1}\right) + \frac{1}{2}\left[1 - \left(\frac{n-2}{n}\right)^m\right]\left(\frac{1}{n+1}\right) \\
&= \frac{1}{n+1} + \frac{1}{2}\left[1 - \left(\frac{n-2}{n}\right)^m\right]\left(\frac{1}{n+1}\right) \\
&= \left(\frac{1}{2}\right)\left(\frac{1}{n+1}\right)\left[3 - \left(\frac{n-2}{n}\right)^m\right]
\end{aligned} \tag{4}
$$

Note that when $m = 1$, we obtain $\mathbb{E}[P_E] = \frac{1}{n}$, indicating an unbiased prevalence estimate. However, we are interested in when $m \geq 2$. Consider when $m = 2$. After some algebra, we obtain $\mathbb{E}[P_E] = \frac{n^2+2n-2}{n^2(n+1)}$. To compare to $\frac{1}{n}$, we consider $\left(\frac{1}{n}\right)^{-1}\mathbb{E}[P_E] = \left(\frac{1}{n}\right)^{-1}\frac{n^2+2n-2}{n^2(n+1)} = \frac{n^2+2n-2}{n^2+n}$. When $n > 2$, $\left(\frac{1}{n}\right)^{-1}\mathbb{E}[P_E] > 1$. Thus, when $m = 2$ and $n > 2$, $\mathbb{E}[P_E] > \frac{1}{n}$. Now, consider the partial derivative of $\mathbb{E}[P_E]$ with respect to $m$: $\frac{\partial}{\partial m}\mathbb{E}[P_E] = -\ln\left(\frac{n-2}{n}\right)\left(\frac{1}{2}\right)\left(\frac{1}{n+1}\right)\left(\frac{n-2}{n}\right)^m$. When $n > 2$, $\frac{\partial}{\partial m}\mathbb{E}[P_E] > 0$, so $\mathbb{E}[P_E]$ is increasing with respect to $m$. Combined with the result that $\mathbb{E}[P_E] > \frac{1}{n}$ for $n > 2$ when $m = 2$, we can conclude that $\mathbb{E}[P_E] > \frac{1}{n}$ more generally, (i.e., when $n > 2$ and $m \geq 2$). □

Although this is a simple case, we have again provided an example of a situation where decision trees are biased towards the minority class. It is worth noting that we can infer from the proof of Theorem 1 that the assumption that the predictors are independent is unnecessary for decision trees fit to purity to be biased. In the extreme case where the predictors are perfectly correlated, it is equivalent to having just one predictor, in which case decision trees provide unbiased prevalence estimates. For correlations with absolute value less than one, we can expect a reduction in the bias observed in Theorem 1, but a bias nonetheless.

We can see from (4) that the bias towards the minority class is caused by the times when the tree splits based on a predictor whose minimum or maximum value is associated with the positive observation. When the tree splits on a predictor that does not have this property, $\mathbb{E}[P_E] = \frac{1}{n+1}$, indicating a bias towards the majority class. However, splits that have this property are very attractive to decision trees. In particular, with only one positive observation, the tree will always split on a predictor with this property if one exists. This provides more understanding of the mechanism for a bias in decision trees; some potential splits are biased towards the majority class and some potential splits are biased towards the minority class, but decision trees tend to prefer the latter splits.

It is important to recall that we have modified the standard decision tree algorithm such that all of a tree's splits must be based on the same predictor. It is natural to question if this modification could be the cause of the bias we have found. To ensure that this was not the case, we conducted a simulation study where this restriction on the decision tree was lifted. This simulation study was conducted in Python (version 3.12.7) using the same assumptions about the distributions of the predictors and their relationship with the response as in Theorem 1. We simulated $m = 2$ $\mathrm{Unif}(0, 1)$ random variables as the predictors and used a RandomForestClassifier (from version 1.5.1 of the scikit-learn library; Pedregosa et al., 2011) to fit the decision tree, using the default settings except with only one tree, both predictors considered at each split, and without bootstrapping (i.e., typical settings for a decision tree). Although using a function designed for random forests may seem like an odd choice for fitting a single decision tree, this is consistent with Phelps et al. (2024), who used this function because they found that results changed slightly when using a DecisionTreeClassifier. We varied the number of observations, considering values of 10, 20, 30, 40, and 50. In each case, the first observation was assigned a label of 1 and the rest were assigned a label of 0. Since the observations were independently uniformly distributed, this process is equivalent to uniformly randomly choosing which observation was assigned a positive label. This simulation procedure was performed 500 000 times.

We define three types of trees that we expect to see generated by the algorithm in our simulation. *Type 1* trees are built using a single split, corresponding to the situation where the positive case is associated with an extreme value for one of the predictors. *Type 2* and *Type 3* trees are built using two splits, corresponding to the situation where the positive case is not associated with an extreme value. In Type 2 trees, the splits are based on the same predictor. In Type 3 trees, the splits are based on two different predictors. Examples of each of these trees are shown in Fig. 4.

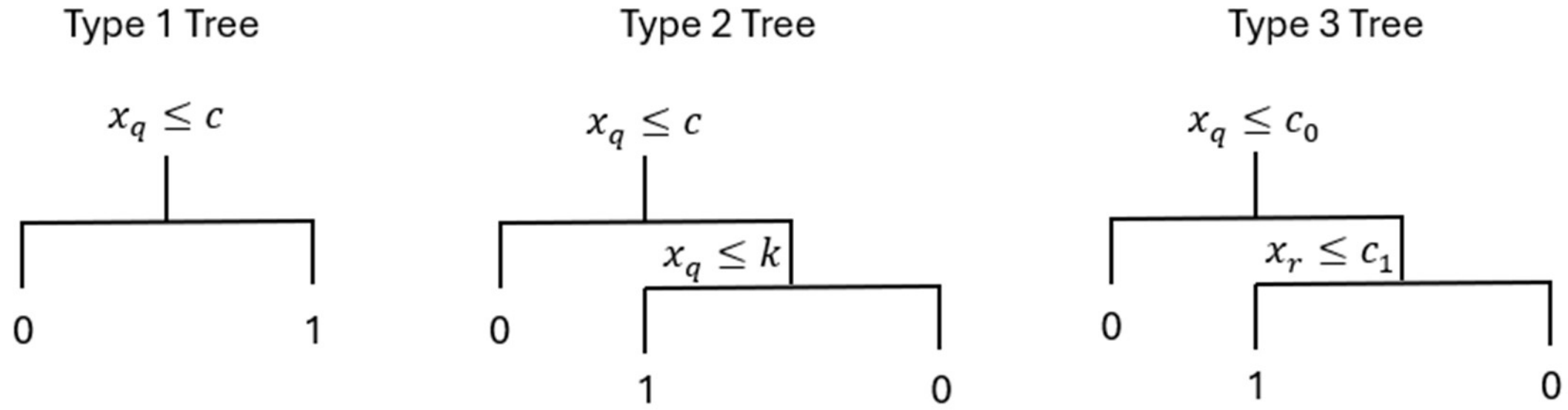

**Fig. 4** Examples of Type 1, Type 2, and Type 3 decision trees. Here, $x_q$ and $x_r$ are predictors with $q \neq r$ and $c$, $k$, $c_0$, and $c_1$ are constants with $k > c$. Note that the leaf labels may occur in a different order depending on the location of the positive instance

For Type 1 trees, we computed the probability of observing such a tree and the ratio of the expected prevalence estimate, $\mathbb{E}[P_E]$, to the true prevalence in the data generating process (i.e., $\frac{1}{n}$). These computations were done by using $m = 2$ in the equations from the proof of Theorem 1. For Type 2 trees, we were able to compute the ratio but not the probability of such a tree, and for Type 3 trees we were unable to compute either value. In Table 1, we provide a summary of our computations as well as the overall ratio of $\mathbb{E}[P_E]$ to the true prevalence, computed under the assumption that Type 3 trees have the same bias as Type 2 trees. This assumption was based on the similar structure of these trees (i.e., both have two splits) and the fact that all Type 3 trees could have been a Type 2 tree (but were not because of an arbitrary decision made by the algorithm), since splitting based on the same predictor twice will also perfectly partition the data in our simulation setting.

**Table 1** Ratios of the expected prevalence estimates to $\frac{1}{n}$. Type 1 trees have only one split and Type 2 trees have two splits, both based on the same predictor. The bracketed values are the expected proportion of Type 1 trees. Type 3 trees, which have two splits but based on different predictors, have been omitted because both their expected prevalence estimate and expected proportion are unknown for this type of tree. The computation of the overall ratio assumes that Type 3 trees have the same bias as Type 2 trees

| $n$ | Type 1 | Type 2 | Overall |
|---|---|---|---|
| 10 | 1.364 (0.360) | 0.909 | 1.073 |
| 20 | 1.429 (0.190) | 0.952 | 1.043 |
| 30 | 1.452 (0.129) | 0.968 | 1.030 |
| 40 | 1.463 (0.098) | 0.976 | 1.023 |
| 50 | 1.471 (0.078) | 0.980 | 1.019 |

The simulation study provided valuable insights to add to our theoretical analysis[2]. First, the simulation study verified the computations shown in Table 1 for Type 1 and Type 2 trees. Second, the simulation study provided insights about the frequency with which we observe Type 2 and Type 3 trees, as well as the bias in Type 3 trees. Notably, Type 3 trees occur with a meaningful frequency and behave very differently from Type 2 trees. Unlike Type 2 trees, Type 3 trees are biased towards the minority class. Therefore, the restriction we imposed on the decision tree actually reduces the bias towards the minority class rather than causing it. We can see by comparing the results in Tables 1 and 2 that the bias in Table 1 is an underestimate of the bias observed when this restriction is lifted; in all five cases, the overprediction more than doubled.

**Table 2** Ratios of the average prevalence estimates to $\frac{1}{n}$, obtained via 500 000 iterations of the simulation. Type 1 trees have only one split, Type 2 trees have two splits of the same predictor, and Type 3 trees have one split based on each of the two predictors. The bracketed values are the observed proportion of each type of tree

| $n$ | Type 1 | Type 2 | Type 3 | Overall |
|---|---|---|---|---|
| 10 | 1.363 (0.361) | 0.910 (0.437) | 1.415 (0.202) | 1.176 |
| 20 | 1.428 (0.191) | 0.953 (0.646) | 1.480 (0.163) | 1.130 |
| 30 | 1.458 (0.129) | 0.967 (0.736) | 1.510 (0.135) | 1.104 |
| 40 | 1.467 (0.098) | 0.975 (0.787) | 1.520 (0.115) | 1.085 |
| 50 | 1.476 (0.079) | 0.981 (0.820) | 1.531 (0.101) | 1.075 |

*3.2.3 The general case: $k$ positive observations*

While the results of Section 3.2.2 provide additional proof that decision trees can be biased towards the minority class, it is an extremely limiting assumption to assume that the dataset has only one positive observation. Thus, in this section, we consider the general case of having $k$ positive observations.

**Theorem 2** *Consider a data generating process that produces datasets of $n$ $(n > 2)$ instances, each with $m$ $(m \geq 2)$ independent predictors and a label that is 0 or 1. Thus, the dataset of predictors, $\boldsymbol{X}$, is $n \times m$. Within each dataset, $k$ $(1 \leq k < \frac{n}{2})$ instances are uniformly randomly assigned label 1 with equal probability, irrespective of the covariate values. Provided $X_j \sim Unif\left(0, u_j\right), u_j > 0,\ \forall j \in 1, 2, \ldots, m$, then a decision tree fit to purity based on only one of the $m$ predictors (chosen by the decision tree algorithm) produces an expected prevalence estimate ($\mathbb{E}[P_E]$) of $\frac{k}{n+1} + \left(\frac{1}{2}\right)\left(\frac{\mathbb{E}[E_{sel}]}{n+1}\right)$, where $E_{sel} \in (0, 1, 2)$ indicates the number of positive*

[2] One surprising finding from our simulation study was that it is possible for the decision tree (produced by the RandomForestClassifier from the scikit-learn library) to have three splits, even though at most two are needed to isolate the positive case. However, this happened infrequently enough that we do not include it in our results.

*cases that are an extreme value with respect to the predictor chosen by the decision tree. This prevalence estimate is biased towards the minority class when* $\mathbb{P}(E_{sel} \geq 1) > \frac{2k}{n}$.

**Proof:** The proof follows a very similar approach to that for Theorem 1, so we omit some details already described there. As before, without loss of generality, we assume $u_j = 1\ \forall j \in 1, 2, \ldots, m$ and consider the length of the region of the selected predictor that predicts a label of 1. We also let $\boldsymbol{X}_s$ be the vector of covariate values from $\boldsymbol{X}$ for the predictor chosen by the decision tree and $S_1, S_2, \ldots, S_{n+1}$ be the spacings on $[0,1]$ between the ordered realizations of $\boldsymbol{X}_s$, including the boundaries. We let $K' \subset \{1, 2, \ldots, n\}$ be the set of all indices of positive cases when the realizations of $\boldsymbol{X}_s$ are sorted in increasing order. Similar to (3), we obtain:

$$P_E = \frac{1}{2}\sum_{i \in K'}(S_i + S_{i+1}) + \frac{1}{2}[I(1 \in K')S_1 + I(n \in K')S_{n+1}] \tag{5}$$

Let $E_{\text{sel}} = I(1 \in K') + I(n \in K') \in (0, 1, 2)$. Then we obtain the following:

$$\begin{aligned}
\mathbb{E}[P_E] &= \frac{1}{2}\left(\frac{2k}{n+1}\right) + \frac{1}{2}\left(\frac{1}{n+1}\right)[\mathbb{P}(1 \in K') + \mathbb{P}(n \in K')] \\
&= \frac{k}{n+1} + \frac{1}{2}\left(\frac{\mathbb{E}[E_{\text{sel}}]}{n+1}\right) \\
&\geq \frac{k}{n+1} + \frac{1}{2}\left(\frac{\mathbb{P}(E_{\text{sel}} \geq 1)}{n+1}\right) \\
&> \frac{k}{n} \text{ when } \mathbb{P}(E_{\text{sel}} \geq 1) > \frac{2k}{n} \quad \square
\end{aligned} \tag{6}$$

Unfortunately, we cannot compute $\mathbb{P}(E_{\text{sel}} \geq 1)$ like we can when $k = 1$, as in Theorem 1. This is because the decision tree is not guaranteed to choose to split on a predictor with a positive case as an extreme value, even when such a predictor exists. However, Theorem 2 does provide further understanding of the way that the bias in decision trees can manifest and a simple way to test if decision trees are biased towards the minority class in this setting. For various values of $k$, $n$, and $p$, we can run simulations to estimate $\mathbb{P}(E_{\text{sel}} \geq 1)$. We can also compute $\mathbb{E}[E_{\text{sel}}]$ and substitute that into (6) to obtain an expected prevalence estimate for each configuration. Example results based on 10 000 simulation runs are shown in Table 3. They show that $\mathbb{P}(E_{\text{sel}} \geq 1) > \frac{2k}{n}$ for every configuration we considered, resulting in prevalence estimates biased towards the minority class.

**Table 3** Estimates of $\mathbb{P}(E_{\text{sel}} \geq 1)$ and the ratio of the expected prevalence estimate, $\mathbb{E}[P_E]$, to the true prevalence, $\frac{k}{n}$, for various values of the number of positive observations in a dataset ($k$), the size of the dataset ($n$), and the number of covariates in the dataset ($m$). Here, $E_{\text{sel}} \in (0, 1, 2)$ indicates the number of positive cases that are an extreme value with respect to the predictor chosen by the decision tree, and if $\mathbb{P}(E_{\text{sel}} \geq 1) > \frac{2k}{n}$, then the decision tree is biased towards the minority class

| $k$ | $n$ | $m$ | $\frac{2k}{n}$ | $\mathbb{P}(E_{\text{sel}} \geq 1)$ | $\frac{\mathbb{E}[P_E]}{k/n}$ |
|---|---|---|---|---|---|
| 5 | 100 | 5 | 0.1 | 0.3932 | 1.0298 |
| 5 | 100 | 10 | 0.1 | 0.6038 | 1.0513 |
| 5 | 100 | 20 | 0.1 | 0.7629 | 1.0672 |
| 10 | 100 | 5 | 0.2 | 0.5463 | 1.0186 |
| 10 | 100 | 10 | 0.2 | 0.6322 | 1.0231 |
| 10 | 100 | 20 | 0.2 | 0.6255 | 1.0231 |
| 50 | 1000 | 10 | 0.1 | 0.5694 | 1.0048 |
| 50 | 1000 | 20 | 0.1 | 0.7161 | 1.0064 |
| 100 | 1000 | 10 | 0.2 | 0.5211 | 1.0018 |
| 100 | 1000 | 20 | 0.2 | 0.5440 | 1.0019 |
| 100 | 10 000 | 15 | 0.02 | 0.2585 | 1.0012 |
| 200 | 1000 | 10 | 0.4 | 0.5258 | 1.0005 |
| 200 | 1000 | 20 | 0.4 | 0.5620 | 1.0006 |
| 1000 | 10 000 | 15 | 0.2 | 0.4763 | 1.0002 |
| 2000 | 10 000 | 15 | 0.4 | 0.5091 | 1.0000 |

Although there is a bias towards the minority class in all configurations, the bias is extremely small in many cases; only the configurations with the smallest dataset size, 100, have prevalence estimates that are more than 1% larger than they should be. This seems to suggest that the bias in decision trees is just a small sample bias, but we do not believe this is the case. Phelps et al. (2024) found that decision trees produced prevalence estimates more than 10% larger than they should be when trained on datasets with 100 000 observations. Recall that we have altered the decision tree algorithm, forcing the tree to make all of its splits based on only one predictor. In Section 3.2.2, we showed that this restriction reduces the bias towards the minority class, and we believe its effect is even stronger in this setting. In Section 3.2.2, at most two splits were needed to isolate the positive case, so restricting the tree to splitting based on just one predictor only influenced at most one split. In the setting with more positive cases, this restriction influences potentially many more splits. Since decision trees are unbiased in this setting when they have only one predictor to choose from, it makes sense that the tree's bias would be reduced when many of its splits are based on choosing from just one predictor. (See Appendix 1 for details showing that $\mathbb{E}[P_E] = \frac{k}{n}$ when $m = 1$.) However, as discussed already, this is not how decision trees work in practice. In practice, decision trees tend to split based on predictors where a minority class observation is associated with an extreme value when the tree has that option (see Table A1 in Appendix 2). We have shown that these splits are biased towards the minority class, so the biases shown in Table 3 should be treated as very conservative lower bounds. Thus, they prove that the bias is towards the minority class, but they are not very effective for understanding the magnitude of the bias. To test this conjecture, we simulated the configurations in Table 3 and fit a decision tree to purity, then made predictions on a testing dataset using that tree. We then compared the prevalence estimates from those predictions to the true prevalence (i.e., $\frac{k}{n}$). This confirmed that the bias is much larger than in Table 3, as the prevalence estimates were all at least 10% larger than the true prevalence (see Table A1 in Appendix 2).

## 4. Discussion

Although there is a widespread and longstanding belief that machine learning models are biased towards the majority class when learning from imbalanced data (e.g., Japkowicz and Stephen, 2002; Guo et al., 2008; Leevy et al., 2018; Megahed et al., 2021), a recent simulation study has drawn that belief into question for decision trees (Phelps et al., 2024). Our work herein provides insight into the biases in decision trees and how they manifest. In Section 3.1, we showed that decision trees can be biased towards either the majority or minority class when the outcome is deterministic given one predictor. In Section 3.2, we considered a stochastic setting and provided a proof that decision trees fit to purity are biased towards the minority class, contrary to popular belief. Although others have investigated biases in decision trees (e.g., Liu et al., 2010; Plante and Radatz, 2024), this is the first theoretical analysis that we are aware of that analytically proves that decision trees fit to purity can produce estimates that are biased towards the minority class.

In order to simplify our analysis and provide a better understanding of decision trees, we have made multiple assumptions. Notably, we have considered cases where the predictors completely explain the outcome and where they have no relation to the outcome at all. While these are extreme cases that we generally would not model in practice, it is important to remember that these assumptions were made to simplify our analysis—not because we believe these conditions are needed for decision trees to be biased towards the minority class. Even in a stochastic setting where the predictors had a relationship with the outcome, Phelps et al. (2024) found that decision trees were biased towards the minority class. We have also altered the way that decision trees learn to simplify our proofs, but we have shown that this simplification actually reduces the bias towards the minority class.

Our findings in Sections 3.2.2 and 3.2.3 both point to a bias towards the minority class, but there is a distinction between them. In Section 3.2.2, our results appear to indicate that the bias decreases as the level of class imbalance increases. In Section 3.2.3, our results appear to show the opposite. The latter results are more in line with Phelps et al. (2024), who found that the bias increased as class imbalance increased, with the exception of one extremely imbalanced case that broke the trend. We posit that the trend observed in Section 3.2.2 may have more to do with increasing the size of the dataset than increasing the level of class imbalance. Between the results of Phelps et al. (2024) and Section 3.3.3, we believe that the bias in decision trees generally increases as class imbalance increases. However, more work is still needed to understand this relationship better.

Our work in Section 3.1 is consistent with the results of Plante and Radatz (2024) in that it indicates that decision trees can be biased in either direction. It may be that the relationships between the predictors and the outcome dictate the direction of this bias, and future work should focus on understanding this better. In Section 3.2, our results are more consistent with the results of Phelps et al. (2024) in that both only found a bias towards the minority class. It is clear from these findings that a bias towards the minority class should be a concern when modeling imbalanced data. Now that we have gained more of an understanding of the bias towards the minority class in decision trees and how it manifests, it is a natural next step to consider how this bias can be reduced or eliminated. Future work is still needed in this area, but we provide several ideas in the following paragraphs.

Based on our findings, one clear way of reducing the bias is to modify decision trees so that they are based on only one predictor. However, fitting trees to only one predictor is very impractical. That said, the datasets we have considered in our analysis need not represent entire datasets. The recursive nature of decision trees means that we might reach a point in the feature space where the problem resembles the entirely stochastic problem considered in Section 3.2. When the tree is focused on a specific subregion of the feature space, our scenario with $\mathbb{E}[Y|\boldsymbol{X}]$ constant over the feature space can approximate any scenario whose conditional mean is sufficiently smooth. Thus, there might be a point in the training process where it makes sense to restrict the decision tree to making its remaining splits based on only one predictor (at least within a specific region of the feature space). However, our results indicate that this will have limited effectiveness and that it certainly cannot eliminate the bias, so other approaches are likely more suitable.

A second way of reducing the bias in decision trees is carefully considering the predictors used in the model. By thoughtfully removing predictors with a limited relationship with the outcome, the bias can be reduced. However, the impact of this approach is minimal, as we found substantial biases even with 10 or fewer predictors (see Table A1 in Appendix 2).

A third method for reducing the bias can be developed by considering the optimal approach in the entirely stochastic setting. In this case, where none of the predictors are related to the outcome, the best thing to do is to not fit a model and just predict a probability of being a positive case of $\frac{k}{n}$ for every observation. If the entire dataset belonged to a single terminal node in the decision tree, this is exactly what the decision tree would do. Thus, the overfitting nature of decision trees fit to purity may be partially to blame for their bias. Therefore, regularization approaches such as pruning, setting a maximum depth for the tree, and/or setting a minimum node size for the tree may be promising approaches for reducing the bias in decision trees. To test the idea of using regularization to mitigate bias in decision trees, we developed a simulation study with 10 predictors, but where only two of those predictors were related to the outcome. Using all 10 predictors, we fit one decision tree to purity and one decision tree to a maximum depth of two. The simulation results confirmed that reducing the depth of the tree can help mitigate the bias towards the minority class (see additional details in Appendix 3).

This idea of using regularization aligns well with other recent studies. Klusowki and Tian (2024) showed that decision trees and random forests are consistent under certain conditions, which may at first seem to contradict our findings. However, although their consistency results involved the tree's depth approaching infinity, they required that $\frac{1}{n} 2^{\text{depth}} \log(n)^2 \log(m) \to 0$ as $n \to \infty$, which is not practical in many settings when the tree is grown to purity. Until recently, fitting decision trees to purity in random forests has generally been considered harmless; the main concern with deep decision trees has been their high variance, but averaging the output of many such trees sufficiently reduces the variance of a random forest. However, Zhou and Mentch (2023) recently showed that random forests can perform better with shallower trees in situations where the signal-to-noise ratio is low. Our work provides an additional reason to question fitting decision trees to purity in random forests. While the process of averaging the outputs across all the decision trees can substantially reduce a random forest's variance, it cannot reduce a systematic bias in the individual trees.

Lastly, another way to effectively reduce the bias in decision trees (without addressing it directly), is to use post-hoc calibration methods, such as the aforementioned Platt's scaling (Platt, 1999) or isotonic regression (e.g., Zadrozny and Elkan, 2002). If using methods like these, it is important to ensure that the calibration method is well-suited to learn the relationship between the original predictions and true probabilities.

## 5. Conclusion

Our work provides theoretical backing to the simulation-based findings of Phelps et al. (2024), showing that decision trees can be biased towards the minority class. To our knowledge, this is the first theoretical analysis to demonstrate such a bias in decision trees. This analysis has provided an understanding of the way in which bias can manifest in decision trees and led to ideas about how to reduce this bias, such as using regularization techniques like setting a maximum tree depth. The implications of our work, however, extend beyond considering which methods should be used but currently are not (e.g., regularization). There are also implications for methods that are currently being used but maybe should not be. It is very common to use methods like undersampling (e.g., Megahed et al., 2021) or cost-sensitive learning (e.g., Chen et al., 2004; Krawczyk et al., 2014) in an attempt to help machine learning models—including tree-based models—learn from imbalanced data because of a belief that these models will neglect the minority class in this setting. Our work suggests that this rationale is not well-founded. There is still a bias, but the bias can instead be in favour of the minority class. Future studies should aim to address if we should still be using these methods when training tree-based models for imbalanced classification problems. It would also be beneficial for future studies to carefully assess the bias in other machine learning models. Our work has specifically focused on decision trees, and our findings do not provide any evidence of a bias towards the minority class for other algorithms, but our work has shown that it should not just be taken for granted that these models are biased towards the majority class.

## Appendix 1

Here, we show that $P_{\mathrm{E}}$ is an unbiased estimator of the true prevalence for the data generating process described in Theorem 2 when $m = 1$. From (6), we have $\mathbb{E}[P_E] = \frac{k}{n+1} + \frac{1}{2}\left(\frac{\mathbb{E}[E_{\mathrm{sel}}]}{n+1}\right)$. Thus, $\mathbb{E}[P_E | m = 1] = \frac{k}{n+1} + \frac{1}{2}\left(\frac{\mathbb{E}[E_{\mathrm{sel}} | m=1]}{n+1}\right)$, and we can compute $\mathbb{E}[E_{\mathrm{sel}} | m = 1]$ as follows:

$$
\begin{aligned}
\mathbb{E}[E_{\mathrm{sel}} | m = 1] &= 2\mathbb{P}(E_{\mathrm{sel}} = 2 | m = 1) + 1\mathbb{P}(E_{\mathrm{sel}} = 1 | m = 1) \\
&= 2\mathbb{P}(E_{\mathrm{sel}} = 2 | m = 1) + 2\mathbb{P}(1 \in K', n \notin K' | m = 1) \\
&= 2\left(\frac{k(k-1)}{n(n-1)}\right) + 2\left(\frac{k(n-k)}{n(n-1)}\right)
\end{aligned} \tag{7}
$$

Substituting (7) into our equation for $\mathbb{E}[P_E | m = 1]$ and simplifying yields $\mathbb{E}[P_E | m = 1] = \frac{k}{n}$.

## Appendix 2

This section provides extended results from the analysis in Section 3.2.3. We compute the probability of at least one predictor having its minimum or maximum value associated with a positive case. This allows us to see how frequently the decision tree chooses such a predictor when one exists. We denote the existence of such a predictor by $E_M$. Using combinatorics, we can compute $\mathbb{P}(E_M)$ as follows:

$$
\mathbb{P}(E_M) = 1 - \left[\frac{\binom{n-2}{k}}{\binom{n}{k}}\right]^m = 1 - \left[\frac{(n-k)(n-k-1)}{n(n-1)}\right]^m \tag{8}
$$

Table A1 also shows an estimate of the ratio of the prevalence estimate to the true prevalence, obtained via simulation. This was added to the table to show the effect of the restriction that the decision tree's splits must all be based on the same predictor.

**Table A1** Estimates of $\mathbb{P}(E_{\mathrm{sel}} \geq 1)$, $\mathbb{P}(E_{\mathrm{M}})$, and the ratio of the prevalence estimate ($P_E$) to the true prevalence $\left(\frac{k}{n}\right)$ for various values of the number of positive observations in a dataset ($k$), the size of the dataset ($n$), and the number of covariates in the dataset ($m$). For reference, we also provide the ratio of the expected prevalence estimate ($\mathbb{E}[P_E]$) to the true prevalence, computed for a tree limited to making all of its splits on one predictor. Here, $E_{\mathrm{sel}} \in (0, 1, 2)$ indicates the number of positive cases that are an extreme value with respect to the predictor chosen by the decision tree, and if $\mathbb{P}(E_{\mathrm{sel}} \geq 1) > \frac{2k}{n}$, then the decision tree is biased towards the minority class. $E_{\mathrm{M}}$ denotes the event where the decision tree has the opportunity to choose a predictor such that $E_{\mathrm{sel}} \geq 1$

| $\boldsymbol{k}$ | $\boldsymbol{n}$ | $\boldsymbol{m}$ | $\mathbb{P}(\boldsymbol{E_{\mathrm{sel}}} \geq \mathbf{1})$ | $\mathbb{P}(\boldsymbol{E_M})$ | $\frac{\boldsymbol{P_E}}{\boldsymbol{k/n}}$ | $\frac{\mathbb{E}[\boldsymbol{P_E}]}{\boldsymbol{k/n}}$ |
|---|---|---|---|---|---|---|
| 5 | 100 | 5 | 0.3932 | 0.4029 | 1.2070 | 1.0298 |
| 5 | 100 | 10 | 0.6038 | 0.6434 | 1.2862 | 1.0513 |
| 5 | 100 | 20 | 0.7629 | 0.8728 | 1.2965 | 1.0672 |
| 10 | 100 | 5 | 0.5463 | 0.6533 | 1.1679 | 1.0186 |
| 10 | 100 | 10 | 0.6322 | 0.8798 | 1.2018 | 1.0231 |
| 10 | 100 | 20 | 0.6255 | 0.9855 | 1.2021 | 1.0231 |
| 50 | 1000 | 10 | 0.5694 | 0.6417 | 1.2381 | 1.0048 |
| 50 | 1000 | 20 | 0.7161 | 0.8716 | 1.2770 | 1.0064 |
| 100 | 1000 | 10 | 0.5211 | 0.8786 | 1.1895 | 1.0018 |
| 100 | 1000 | 20 | 0.5440 | 0.9853 | 1.2038 | 1.0019 |
| 100 | 10 000 | 15 | 0.2585 | 0.2603 | 1.2937 | 1.0012 |
| 200 | 1000 | 10 | 0.5258 | 0.9885 | 1.1071 | 1.0005 |
| 200 | 1000 | 20 | 0.5620 | 0.9999 | 1.1078 | 1.0006 |
| 1000 | 10 000 | 15 | 0.4763 | 0.9576 | 1.1974 | 1.0002 |
| 2000 | 10 000 | 15 | 0.5091 | 0.9988 | 1.1065 | 1.0000 |

## Appendix 3

To assess the effectiveness of limiting the maximum depth of the tree for reducing the bias towards the minority class, we developed a simulation. The simulation is based on one of the simulations in Phelps et al. (2024), involving the same 10 uniformly distributed predictors (see Table A2). However, it differs from their simulation in that only two of the predictors are related to the outcome. When the fifth and sixth predictors are greater than 3.75 and 2.25, respectively, the outcome takes value 1 with probability 0.9820. This occurs in 6.25% of cases. In all other cases, the outcome is 1 with probability only 0.1192, resulting in an overall prevalence of 0.1731. We then fit two decision trees using all 10 predictors to a training dataset of 500 observations and compared the predictions of the trees to the true probabilities in a testing dataset. The difference between the decision trees is that one was fit to purity and one was fit to a maximum depth of two, which is the depth needed to properly characterize this data generating process. We repeated this process 10 000 times to obtain a precise estimate of the bias in each decision tree. For the tree fit to purity, we obtained a ratio of 1.1019 between the predictions and true probabilities, indicative of the bias towards the minority class that is expected based on our other results. For the tree fit to a maximum depth of two, we obtained a ratio of 0.9917, indicating that the bias towards the minority class was eliminated.

**Table A2** The minimum and maximum value for each of the 10 uniformly distributed predictors in the simulated datasets

| Covariate | Minimum | Maximum |
|---|---|---|
| 1 | -0.4 | 0.6 |
| 2 | -0.2 | 0.8 |
| 3 | -0.4 | 1 |
| 4 | -0.1 | 0.9 |
| 5 | 0 | 5 |
| 6 | 0 | 3 |
| 7 | 1 | 4 |
| 8 | 1 | 7 |
| 9 | 1 | 3 |
| 10 | 0 | 2 |